# A criterion for Artificial General Intelligence: hypothetic-deductive reasoning, tested on ChatGPT

Louis Vervoort*, Vitaliy Mizyakov, Anastasia Ugleva

*Higher School of Economics, Moscow*

**email for correspondence: lvervoort@hse.ru*

20.06.2026

**Abstract.** We argue that a key reasoning skill that any advanced AI, say GPT-4, should master in order to qualify as 'thinking machine', or AGI, is hypothetic-deductive reasoning. Problem-solving or question-answering can quite generally be construed as involving two steps: *hypothesizing* that a certain set of hypotheses T applies to the problem or question at hand, and *deducing* the solution or answer from T – hence the term hypothetic-deductive reasoning. An elementary proxy of hypothetic-deductive reasoning is causal reasoning. We propose tests for both types of thinking, and apply them to ChatGPT. Our study suggests that, at present, the chatbot has a limited capacity for either type of reasoning, as soon as the problems considered are somewhat complex. However, we submit that if an AI would be capable of this type of reasoning in a sufficiently wide range of contexts, it would be an AGI.

## 1. Introduction.

Large Language Models (LLMs) as ChatGPT[1] have recently made disruptive progress in answering questions and solving problems – thus mimicking human thinking in ever more realms. Will this progress ever stop, or will AI and machine learning (ML) / deep learning (DL) become superior to humans in all cognitive tasks? Clearly, in some domains AI already outperforms humans; in some fields it is still much weaker. But what are the key types of reasoning or thinking that characterize intelligent humans, and that pose the largest challenge to advanced AI, including LLMs? ChatGPT, both based on GPT-3.5 and GPT-4, has already been tested on a wide series of different reasoning skills, sometimes with truly impressive, and to some extent unexpected, results. A partial list of examples includes logical reasoning (e.g. Liu et al. 2023, Bang et al. 2023), mathematical reasoning (e.g. Frieder et al. 2023, Wardat et al. 2023), physical reasoning (e.g. Lehnert 2023, West 2023), psychological reasoning (e.g. Hagendorff

[1] From the company OpenAI (https://openai.com/). ChatGPT, based on the Generative Pre-trained Transformer (GPT) language model (versions GPT-3.5 and GPT-4), is freely available with GPT-3.5, less powerful than GPT-4 (cf. OpenAI 2023).

2023, Holterman and van Deemter 2023), medical reasoning (e.g. Bhayana et al. 2023) and several other types; this list may well grow exponentially in the near future (see also e.g. Bang et al. 2023, Borji 2023, Bubeck et al. 2023, Espejel et al. 2023, Huang and Chang 2022, Mahowald et al. 2023). The unprecedented capabilities of LLMs trigger a range of fundamental questions as to the nature of cognition and reasoning, questions relevant in computer science, epistemology, philosophy of mind, cognitive science, psychology and several other disciplines. One such question is whether a general scheme of reasoning exists that could subsume (a significant part of) all types of reasoning. This question is also important for practical reasons: if a technology would be developed that allows this subsuming scheme to successfully work in one knowledge-field, it could then work for others too. An indirect goal of this article is to incite philosophers to participate in the pressing debates that emerge in the wake of a technological revolution that has an enormous potential for societal impact, besides such obvious connections to ancient philosophical questions (for an early discussion, cf. Arkoudas 2023).

Clearly, regarding brute 'knowledge amassment' – amassment of documented facts and passive knowledge – many types of AI are insuperable due to their access to gigantic memory storage. But humans are capable of 'genuine' reasoning, 'genuine' understanding, by actively applying this passive knowledge to solve problems[2]. Drawing on concepts of philosophy of science (notably Hempel 1965, Laudan 1977, Goldman 1983, Thagard 1988, Vervoort et al. 2021, Vervoort et al. 2023), we propose here a candidate for such a subsuming scheme of reasoning. We argue that, from a logical or epistemological point of view, the key cognitive capacity for general problem solving is 'hypothetic-deductive' or 'theory-based' reasoning – following (Vervoort et al. 2021), we use a new umbrella term that best captures our findings. This is essentially the reasoning that is used in science (Hempel 1965); but we argue here that 'common-sense' and other types of thinking follow largely the same scheme. A related cognitive capacity is causal reasoning, which is vital for human survival from our very first days on earth: we need to be able to identify cause-effect relations in order to guide our lives in all (new) situations encountered. From this respect, it seems not surprising that infants can identify causal relata from very early on (Leslie and Keeble 1987). Causal reasoning and hypothetic-deductive reasoning are closely related (e.g. Woodward and Ross 2021), as recalled in the following.

[2] We adopt the pragmatic conception of intelligence, centered on problem solving, which has become popular in the (philosophy of) AI since Turing (cf. Norvig and Russell 2015, Bringsjord and Govindarajulu 2018). In this perspective the distinction between 'mastering a type of reasoning' and 'emulating/simulating this type of reasoning' largely disappears: what counts is the capacity to solve problems. Furthermore, for our practical purposes, we will not delve into the links and possible differences between schemes of reasoning, explanation and understanding (for more on this topic, see e.g. Woodward and Ross 2021, and references therein).

How good is ChatGPT at hypothetic-deductive and causal reasoning – or at least at emulating/simulating such reasoning? In view of the imminent development of ever more powerful AIs, it is important to have practical tests and subsuming concepts that allow one to easily compare different AIs as to their general reasoning skills – also for instance for ethical reasons. We propose here simple tests for the two mentioned cognitive capacities. In contrast to computer scientists, who have developed comprehensive benchmark tests (e.g. BIG-bench, Srivastava et al. 2022), our goal is to motivate and present test *methods*, rather than providing comprehensive benchmarks and test results. However we illustrate the methods with some test results on ChatGPT (based on GPT-4), enough for supporting our conclusions, especially when coupled to other published results (e.g. Bang et al. 2023, Espejel et al. 2023, Liu et al. 2023, Huang and Chang 2022). (Our test for abstract causal reasoning could easily be developed and formalized, but that is not our goal here.) For the experiments we first used the free version of ChatGPT based on GPT-3.5 (in April-May 2023) and then (in July 2023) the GPT-4 based version, commercialised as 'ChatGPT Plus'. In the following we refer to these versions as ChatGPT(3) and ChatGPT(4), respectively.

We thus report first results showing that, as of the time of testing, ChatGPT has limited capacities for causal reasoning and hypothetic-deductive reasoning, as soon as the problem context is slightly sophisticated. Depending on one's perspective, this is surprising, or not. This might not be surprising, if LLMs are 'simply' based on statistical text completion, and have no comprehensive 'world model' (OpenAI 2023). On the other hand, in view of the impressive results obtained with ChatGPT, for instance in coding and mathematical creativity (e.g. Bubeck et al. 2023), many recent reports are more optimistic, even as to the qualification of ChatGPT as an AGI (artificial general intelligence): see e.g. the verdict of a team of researchers from Microsoft (Bubeck et al. 2023). These authors state: "A question that might be lingering on many readers' mind is whether GPT-4 truly understands all these concepts, or whether it just became much better than previous models at improvising on the fly, without any real or deep understanding. We hope that after reading this paper the question should almost flip, and that one might be left wondering how much more there is to true understanding than on-the-fly improvisation" (Bubeck et al. 2023). Let us try to answer this question.

LLMs as ChatGPT and AI in general will in the future surely be enhanced with new computational tools, multiplying their capacities[3]. Also for these enhanced tools, we maintain our conclusion: in order to qualify as genuine and trustworthy 'thinking machines', or AGI, these AI-systems should be able to master hypothetic-deductive and causal reasoning, and pass tests similar to ours at expert level. Let us emphasize that trustworthy AI is explainable AI (Doshi-Velez and Kim 2017, Gilpin et al. 2019, Arrieta et al. 2020), in other words, AI that can answer 'explanation-seeking why-questions'; still in other words, AI that is capable of hypothetic-deductive reasoning.

[3] For a recent and notable example, cf. (Wolfram 2023).

For our causal reasoning test, we use and elaborate on a compendium of causal problems that have been studied in the literature (Paul and Hall 2013), and that provides a catalogue of abstract causal situations in the form of 'neuron diagrams'. These neuron diagrams can be transcribed and questions about them submitted to ChatGPT. Our test focuses on abstract causal reasoning, to be distinguished from causal reasoning applied to concrete contexts, as it was tested in (Bang et al. 2023, Tu et al. 2023); it seems no surprise that ChatGPT performs better on the latter task. We have studied ChatGPT's hypothetic-deductive reasoning skills mostly in the context of elementary physics. Testing its capacity to answer physics questions has recently been done (Kortemeyer 2023, West 2023, Holmes et al. 2023). But answering physics questions correctly can be done by skilfully copying from internet resources without having genuine understanding. Here we motivate the use of a method we call 'hypotheses listing', which allows for an immediate insight into the (simulated/emulated) reasoning skills. This method is a variant of 'chain-of-thought prompting', already studied e.g. in (Wei et al. 2022, Kojima et al. 2022, Chen et al. 2023).

The article is organised as follows. In Section 2 we define hypothetic-deductive reasoning and argue that it can be considered as a subsuming scheme of rational thinking and understanding. First we focus on scientific thinking, postponing a generalisation to common-sense thinking to Section 5. We present our view on how hypothetic-deductive reasoning is related to causal reasoning, and explain why causal reasoning can be considered as a simplified proxy, or rather as a simplified derivative of hypothetic-deductive reasoning. In Section 3 we propose a test for abstract causal reasoning, and show results obtained with ChatGPT(4). As an immediate application of our model, we propose in Section 4 a test scheme for hypothetic-deductive reasoning, and use it to assess ChatGPT. We illustrate the proposed method with examples. In Section 5 we discuss the scope of these results from the perspective of developing AGI – the goal of, for instance, the company that commercialises ChatGPT. Finally, Section 6 concludes with a general criterion for defining an AI as an AGI.

## 2. Hypothetic-deductive reasoning, its relevance and link with causal reasoning.

From an epistemological or logical point of view, how are humans capable of genuine reasoning and genuine understanding? How can we use or apply stored passive knowledge in order to solve new problems, simple or sophisticated? To answer this question, many philosophers have found it instrumental to look at science, in particular physics: in the scientific realm the logical scheme of problem-solving and explanation has extensively been studied (e.g. Hempel 1965, Laudan 1977, Goldman 1983, Thagard 1988, Woodward and Ross 2021). In a sense, the 'problem spaces' in physics are much simpler than those of everyday life. Physics deals with the 'simplest' systems in that they are usually isolated model entities, such as a ball, an atom, a laser beam, and, especially, in that they are inanimate. In everyday life our thinking often concerns incommensurably more complex systems – for instance humans, having free-willed consciousness, behaving largely unpredictably. Now, in physics it is clear how questions are answered and problems solved: in a nutshell, *by applying the relevant theory to the question/problem at hand*. (We will from now on not distinguish anymore between 'questions' and 'problems': they are dealt with in the same way (Goldman 1983). Also, we look at rational or intellectual problems that can be made textually explicit, for instance as they are asked in physics exams, or encountered in research, or asked to ChatGPT. But perhaps some problems might need other problem-solving methods than the general theory-based method we consider here, e.g. using intuitive know-how rather than explicit knowing-that.)

As an example, let us look at a concrete real-world question as "why does this gas have a pressure of 0.03 bar?" Physicists consider a problem solved if they have shown that the answer deductively follows from the relevant theory, for instance thermodynamics, mechanics, electrodynamics etc. For the example, they can use thermodynamics, containing the gas law, to deduce following answer: this gas is enclosed in a volume V and has temperature T (as can be measured), and it follows from the gas law that its pressure therefore is n.R.T/V (here n and R are known constants). If this calculation leads to a number that is in sufficient approximation equal to the measured pressure of 0.03 bar, then the question is answered.

Note that this explanation involves two essential steps: *hypothesizing* that a certain theory applies to a certain problem; then *deducing* that the answer logically follows from this theory, i.e. from this (collection of) law(s) (and particular facts). This is the *hypothetic-deductive reasoning* that is the hallmark of physical problem solving. (In any case if the problem space has been explored, so that adequate theories are known. The vast majority of problems that physicists tackle is of this kind – problems of Kuhn's 'normal science'. Only a relatively small number of questions, at the forefront of science, need the construction of new theories and are therefore at best only speculatively answered. But the speculative answers nevertheless follow the above hypothetic-deductive scheme, involving now speculative theories/laws.)

So in physics the basic logical scheme of problem solving is quite clear. We leave certain aspects undiscussed, but we only wish to identify the essential ingredient of scientific reasoning that can be compared to the capacities of the best contemporary AI. In philosophy of science, this scheme is usually called the 'deductive-nomological' or 'covering law' model of scientific explanation (Hempel 1965, Woodward and Ross 2021), formalized notably by Hempel, but we prefer '*hypothetic-deductive'* or *'theory-based' reasoning*: both terms are more illustrative of what we mean[4]. The term 'hypothetic-deductive' reasoning aptly emphasizes that this type of thinking involves *two* processes, both, a priori, of daunting difficulty for AI. For our purposes it suffices to consider theories as coherent systems of hypotheses (laws, regularities, generalisations, norms,…).

It is well known that also in other natural and social sciences hypothetic-deductive reasoning (as we defined it) is ubiquitously used for problem solving. In these disciplines theories may not be as well-defined as in physics, but answering questions is usually also based on starting from coherent sets of hypotheses, involving laws, regularities, generalisations, norms. Elsewhere it is argued that this is the general framework for rational thinking and rational[5] understanding, not only in science but also in the context of every-day life (Vervoort et al. 2021). We delve slightly deeper into this argument in Section 5, but we can stick here to the weaker claim that hypothetic-deductive reasoning is one key ingredient of rational thinking, more precisely a *necessary condition*. Then, if AGI wants to become reality, it will have to master or emulate this type of reasoning.

It makes sense to test ChatGPT first with a simplified proxy of hypothetic-deductive reasoning, namely causal reasoning. The basic task of causal reasoning is identifying causes (as opposed to effects) in a given causal situation. Humans have an amazing intuitive capacity to identify causes (Leslie and Keeble 1987) – a result confirmed by our study. In certain somewhat subtle cases philosophers debate about how exactly to define the concept of cause, but in practice people usually have converging intuitions about *what causes what*. For our purposes, it suffices to adopt the popular counterfactual interpretation of cause: in essence, C is a cause of effect E if the following holds: if C would not occur, E would not occur

[4] Hypothetic-deductive thinking should not be confused with the 'hypothetico-deductive method' of scientific discovery: the method that is based on formulating hypotheses from which testable deductions are derived. This is a type of scientific practice à la Popper, not a reasoning scheme.

[5] To be opposed to, for instance, intuitive understanding or understanding-by-doing, which does not need to be rationalized by explicit reasons.

(Lewis 1973, Paul and Hall 2013). For instance, in the above example, one might say that the fact that the gas has a temperature T causes it to have pressure P, in short, T causes P. This fits to our definition: if the temperature would *not* be given by T, the pressure would *not* be given by P.

Theories contain of course causal information (among other things); it is then no surprise that there is a link between theory-based and causal reasoning – as is well known. Quite generally, one needs a theory to extract causes in a given situation. For instance, to elucidate who is the culprit in a murder case, one better have a solid theory about it. In order to identify the cause of the earth's movement around the sun – gravity –, one needs the theory of mechanics of Newton. In somewhat more detail, from a logical or epistemological point of view, theories and laws allow for identifying what is 'necessary' for something to occur; that is why they are handy tools for identifying causes. Indeed, thanks to theories one can imagine what would happen if one 'takes out' or alters some phenomenon (represented by a variable in formalized theories), and see whether other phenomena change accordingly – in other words, study the influence of causes on effects, in agreement with the counterfactual interpretation of cause just mentioned (more in Vervoort 2013). Of course, theories often do much more work than only identify causes: e.g. in quantitative science they allow for making precise, quantitative predictions. Thus, the capacity to identify causes can be seen as a much simplified proxy of hypothetic-deductive reasoning, or more precisely, as an elementary application or derivative of theory-based reasoning[6]. One would presumably not say that humans master a theory if they cannot draw causal information from it.

In the next Section we devise a test for causal reasoning, and show the test results obtained with ChatGPT(4).

## 3. Test for causal reasoning, and results obtained by ChatGPT.

In order to test the causal reasoning ability of ChatGPT, we ask it to identify the causes in 'neuron diagrams', often used to study causation. Such diagrams allow to represent a wide variety of causal situations as well as problems discussed in the literature, including causal redundancy, omissions, and transitivity. A representative case is given in Fig. 1, taken from the reference work by Paul and Hall (2013).

[6] Note that it is only from the logical-epistemic point of view that causal reasoning is a simplified derivative of theory-based reasoning. Of course, infants, who can intuitively and implicitly master causal reasoning, do not use explicit theories to identify causes. Presumably they learn cause-effects relations by frequent observation of similar-enough instances.

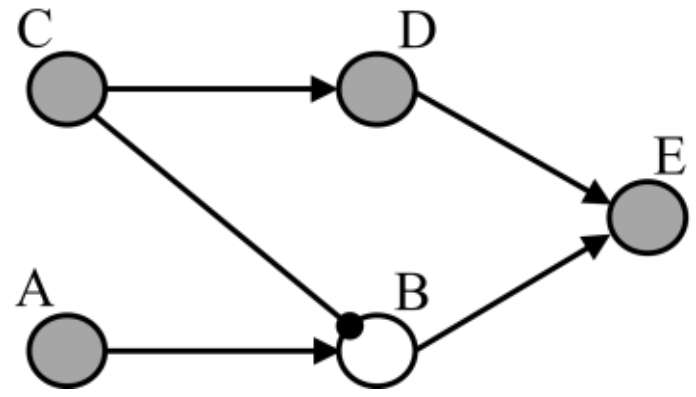

Fig. 1. Typical neuron diagram (reproduced from Fig. 1, Paul and Hall 2013, p. ix), explained in the text. 'Firing' or 'on' neurons are shaded.

In the diagram of Fig. 1 five subsystems, here represented by five neurons, are in causal interaction in the following way. Neuron C emits a stimulating signal (represented by an arrow) towards neuron D, causing it to fire; D likewise causes E to fire (all firing/'on' neurons are shaded). A stimulates B but B is inhibited to fire due to the inhibitory signal from C, represented by a line with a blob. The temporal order goes from left to right: C and A fire at t1, B and D react at t2, E reacts at t3, where t1 < t2 < t3. Looking at Fig. 1, most people would effortlessly state that the cause(s) of E's firing are the firing of C and/or D. Or if they wish to be more precise, they could say that C is the 'root' cause at t1 and D the 'proximate' cause at t2. Intuitively the firing of A would *not* be identified as a cause, because its action is blocked by C. A series of 26 diagrams is given in Table 1 in Appendix A. These include a sizeable part of the cases discussed in (Paul and Hall 2013), as well as several variations of these diagrams.

Even if causal intuitions of people largely overlap – there is little discussion about what causes what in most diagrams presented in Table 1 – it is not trivial to precisely define the concept of cause; in subtle cases the definition is debated. As mentioned above, one widely accepted strategy to identify causes of Y is to look at what happens if some event X would not occur (ceteris paribus, i.e. all unrelated events kept fixed); if then Y does not occur X is a cause (if X also occurs before Y). As we show in Appendix B, this method, if slightly adapted, works quite well in all the cases we studied. It works in particular to identify proximate causes, i.e. causes of effects that are *only one step later* in time in the diagrams. For instance, in Fig. 1, this method immediately allows one to find that 'D+' (the firing of D) is a cause at t2 of E+ at t3, and that B- (the non-firing of B) is *not* a cause of E+. So proximate causes are easily identified by the simple counterfactual rule. But root causes, i.e. causes occurring at t1 in the diagrams, are sometimes less easily identified. To make the counterfactual method fully analytically precise for all situations depicted in Table 1 some specifications are needed; the analytic definition we used is given in Appendix B. Now, such a detailed analytic definition is in principle not even necessary to test ChatGPT at the present stage: wrong answers can very easily be spotted.

We asked ChatGPT(3) and ChatGPT(4) questions based on the neuron diagrams gathered in Table 1 (Appendix A). To this end, the diagrams must first be transcribed in text. This is a straightforward procedure, as shown by the examples given in Table 2. Then we always ask essentially the same question, namely whether the last neuron in the diagram fires or not, and what the causes of the firing or not firing are. Table 2 also shows the correct answers and the answers given by ChatGPT(4).

| Diagram | Transcription + Question | Correct answer | ChatGPT(4)'s answer |
|---|---|---|---|
| **1. (From PH, Fig. 1, p. ix)** | Suppose time t1 is earlier than time t2, which is earlier than time t3. If C would occur at t1, D would occur at t2. If D would occur at t2, E would occur at t3. If A would occur at t1, B would occur at t2, unless C would occur at t1. If B would occur at t2, E would occur at t3. Suppose C and A occur at t1. Does E occur at t3? What is/are the cause(s) of E's occurring or not occurring? | Yes.<br>**C+(t1); D+(t2)** | If both A and C occur at t1, then according to your stipulations, at t2, D would occur (caused by C) and B would not occur (because A at t1 prevents B from occurring at t2). Since B does not occur at t2, the condition for E to occur at t3 based on B is not met. However, D does occur at t2, which, according to your conditions, does cause E to occur at t3. So, E does occur at t3, and the cause is D occurring at t2, which in turn is caused by C occurring at t1.<br>***<CORRECT>*** |
| **2.** | Suppose time t1 is earlier than time t2, which is earlier than time t3. If C would occur at t1, D would occur at t2. If D would occur at t2, E would occur at t3 unless B would occur at t2. If A would occur at t1, B would occur at t2, unless C would occur at t1. Suppose C and A occur at t1. Does E occur at t3? What is/are the cause(s) of E's occurring or not occurring? | Yes.<br>**C+(t1); D+(t2), B-(t2)** | Given the conditions and that C and A occur at t1, we can deduce the following:<br>1. Since C occurs at t1, D will occur at t2.<br>2. The occurrence of C at t1 means that even though A also occurs at t1, B will not occur at t2 (as per the rule "If A would occur at t1, B would occur at t2, unless C would occur at t1").<br>3. Since D occurs at t2 and B does not, there is nothing to prevent E from occurring at t3. Therefore, E does occur at t3.<br>The cause(s) of E's occurring is the presence of C and A at t1 which leads to the presence of D at t2, and the non-occurrence of B at t2 (due to the presence of C at t1), allowing E to occur at t3.<br>***<PARTLY CORRECT>*** |
| **5. (From PH, Fig. 10, p. xi)** | Suppose time t1 is earlier than time t2, which is earlier than time t3, which is earlier than time t4. If A would occur at t1, A1 would occur at t2. If A1 would occur at t2, A2 would occur at t3. If A2 would occur at t3, E would occur at t4, unless F occurs at t3. If C would occur at t1, B and D would occur at t2. If B would occur at t2, F would occur at t3, unless D would occur at t2. Suppose both C and A occur at t1. Does E occur at t4? What is/are the cause(s) of E's occurring or not occurring? | Yes.<br>**A+(t1); D+(t2); F-(t3)** | Given the conditions, E does not occur at t4.<br>Here's how the events unfold:<br>1. At t1, both A and C occur.<br>2. At t2, as a result of A's occurrence at t1, A1 occurs. As a result of C's occurrence at t1, both B and D occur.<br>3. At t3, A2 occurs because A1 occurred at t2. However, F does not occur because although B occurred at t2, D also occurred at t2, which prevents F from occurring.<br>4. At t4, E does not occur because even though A2 occurred at t3, F did not occur at t3.<br>So, the primary reason for E not occurring at t4 is the non-occurrence of F at t3, which is due to the occurrence of both B and D at t2, which in turn was caused by C's occurrence at t1.<br>***<INCORRECT>*** |
| **10. (From PH, Fig. 14, p. xiii)** | Suppose time t1 is earlier than time t2, which is earlier than time t3. If A would occur at t1, B would occur at t2. If A would occur at t1, D would occur at t2, unless C would occur at t1. If B would | Yes.<br>**A+(t1), C+(t1); B+(t2), F+(t2)** | If both A and C occur at t1, then at t2, according to your stipulations:<br>1. B would occur because A has occurred at t1.<br>2. D would not occur because C occurring at t1 prevents D from occurring at t2. |
| 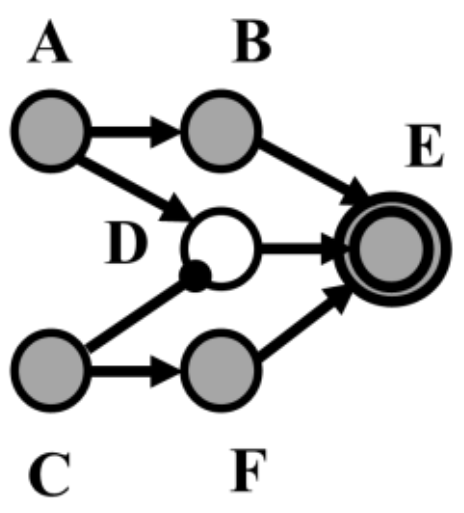 | occur at t2, E would occur at t3, if also either D at t2 or F at t2 (or both) would occur. If C would occur at t1, F would occur at t2, and D would not occur at t2. If F would occur at t2, E would occur at t3 if also either D or B (or both) would occur at t2. If D would occur at t2, E would occur at t3 if also either B or F (or both) would occur at t2. Suppose both A and C occur at t1. Does E occur at t3? What is/are the cause(s) of E's occurring or not occurring? | | 3. F would occur because C has occurred at t1.<br>Given these outcomes, we then look at the conditions for E occurring at t3. You stated that E will occur at t3 if B, D, or F (or any combination of them) occur at t2.<br>In this situation, both B and F occur at t2. Hence, even without D occurring at t2, the condition for E to occur at t3 is satisfied.<br>Therefore, E does occur at t3. The causes of E's occurrence are B and F occurring at t2, which were respectively caused by A's occurrence at t1 and C's occurrence at t1.<br>***<CORRECT>*** |
| **18. (From PH, Fig. 44, p. xxi)**<br>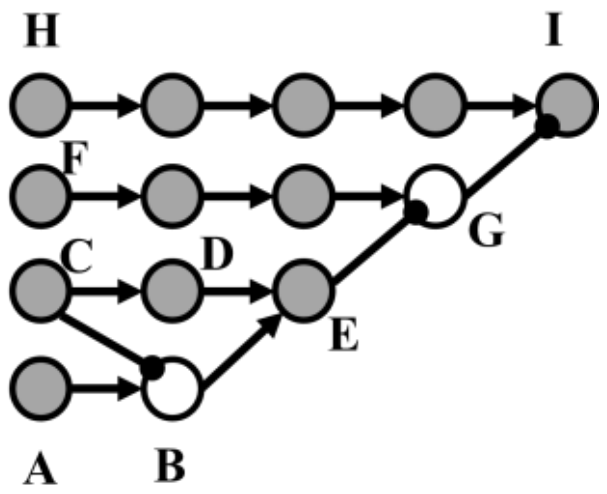 | Suppose time t1 is earlier than time t2, which is earlier than time t3, which is earlier than time t4, which is earlier than time t5. If H would occur at t1, H1 would occur at t2. If H1 would occur at t2, H2 would occur at t3. If H2 would occur at t3, H3 would occur at t4. If H3 would occur at t4, I would occur at t5, unless G would occur at an earlier time. If F would occur at t1, F1 would occur at t2. If F1 would occur at t2, F2 would occur at t3. If F2 would occur at t3, G would occur at t4, unless E would occur at an earlier time. If C would occur at t1, D would occur at t2. If D would occur at t2, E would occur at t3. If A would occur at t1, B would occur at t2, unless C would occur at an earlier time. If B would occur at t2, E would occur at t3. Suppose H, F, C and A occur at t1. Does I occur at t5? What is/are the cause(s) of I's occurring or not occurring? | Yes.<br>**H+(t1), C+(t1); H3+(t4), G-(t4); etc.** | Given the conditions and the events at t1 (H, F, C, A), let's follow the chain of events:<br>1. H occurs at t1, so H1 will occur at t2. Then, H1 leads to H2 at t3, H2 to H3 at t4, and without interference, H3 would lead to I at t5.<br>2. F occurs at t1, so F1 will occur at t2. Then, F1 leads to F2 at t3, and without interference, F2 would lead to G at t4.<br>3. C occurs at t1, so D will occur at t2. Then, D leads to E at t3.<br>4. A occurs at t1. But since C also occurs at t1, B will not occur at t2 (as per the rule "If A would occur at t1, B would occur at t2, unless C would occur at t1"), so there's no additional path to E from here.<br>Following the sequence, the conditions indicate that G should occur at t4 (as F2 is expected at t3 and there's no E prior to t3 to prevent it). Since G happens at an earlier time than t5, this will prevent I from occurring at t5.<br>So, I does not occur at t5. The cause of I's non-occurrence is the event sequence initiated by F at t1, which leads to G occurring at t4 (before t5). Without the occurrence of F at t1, the sequence leading to G (and thus preventing I) wouldn't happen.<br>***<INCORRECT>*** |

**Table 2**. ChatGPT(4)'s answers (column 4) to causal questions based on neuron diagrams (see full table in Appendix A). Column 2 gives the transcriptions and questions and column 3 the correct answers (here C+(t1) stands for 'the firing/occurrence of C at time t1'; B-(t2) for the 'the non-firing/non-occurrence of B at t2'; etc.). In the last column, incorrect parts are underlined. All diagrams except 2 are discussed in (Paul and Hall 2013) (PH in the table). In diagram 10, the last neuron has a double border, meaning that it only fires upon reception of at least two stimulating signals. In diagram 5, the intermediate neurons A1, A2 between A and E are not labelled for simplicity; same simplification in diagram 18.

The 'correct answers' columns in Table 1 and 2 collect the intuitive causes, in any case those identified with DEF1 (Appendix B). It is perhaps not always easy or immediate to identify *all* the causes of an event in a neuron diagram. But we emphasize that for our test this is not even necessary, since it is easy to ascertain that an answer is wrong. As said above, the proximate causes are most easily identified, both intuitively and (hence) by a simple counterfactual rule. Sometimes the initial or root causes are somewhat less easily identified; but then one can use DEF1. (On a speculative note, our research leads us to surmise that humans intuitively, implicitly use something as DEF1 to hunt for causes.) In Tables 1 and 2 we have in most cases only indicated root and proximate causes, sometimes also intermediate causes, notably when confusion could arise.

The results we obtained were the following. For all diagrams except 5 and 18, ChatGPT(4) correctly answered the question of whether the last neuron fires or not, a much better result than obtained with ChatGPT(3) (see below). However, for the crucial part of the test, focusing on finding causes, the results were less convincing. As indicated in Table 1 (Appendix A), ChatGPT(4) only correctly indicated all causes for 13 of the 26 submitted diagrams; for 2 more diagrams the answer was partly correct. Examples of the full answers can be found in Table 2. One will notice that these often seem to make perfect sense, at least on the surface. But identifying causes is, in principle, an all-or-nothing task: overlooking causes can have dramatic consequences, as any engineer, scientist, health care worker etc. can tell. In Tables 1 and 2 our evaluation <(IN)CORRECT> refers to the question as a whole, in particular the identification of causes. We emphasize that for humans finding causes in diagrams as the above is often simple, at least with a minimal training.

ChatGPT(4)'s score is much better than the one obtained by ChatGPT(3), to which we had submitted cases/diagrams 1 until 21 (on 08.05.2023). ChatGPT(3) found for only one diagram (20) the correct causes. This significant increase in capability due the parametric upscaling in GPT-4 is generally seen in tests. As a side-note, the diagrams 1 to 21 are either discussed in (Paul and Hall 2013), or immediate variants; it is also on these diagrams, submitted to ChatGPT(3) in May 2023, that ChatGPT(4) scores best. When we submitted in July more complex diagrams as 22-26, not discussed in (Paul and Hall 2013), ChatGPT(4)'s answers were all incorrect (cf. Table 1). However, we didn't perform a systematic study of ChatGPT's score as a function of diagram complexity; and of course, with further upscaling of the LLM these scores must be expected to improve.

As another general remark, and as is illustrated in the last column of Table 2, ChatGPT's answers showed problems of sequential logical reasoning: intermediate results are not used for subsequent deductions. This point is confirmed by the hypothetic-deductive reasoning test presented in the next Section, and is known from the literature (e.g. Bang et al. 2023, Espejel et al. 2023, Huang and Chang 2022).

We conclude from above tests that ChatGPT(3) and ChatGPT(4) have, at the time of writing, a limited capacity for abstract causal reasoning, where humans normally can perform better and with little effort.

## 4. Test for hypothetic-deductive reasoning, and results obtained by ChatGPT.

As argued in Section 2, causal reasoning can be considered a simplified derivative of hypothetic-deductive reasoning. Since ChatGPT performs only moderately well on causal reasoning, we expected that it would not be too difficult to find instances of failing hypothetic-deductive reasoning, cases where (minimally trained) humans would not fail. This is confirmed by our tests, of which we will now present some results. We present here only a few examples to illustrate the concepts and method proposed. Other authors have studied ChatGPT's step by step reasoning capacities in specific disciplines (e.g. Bang et al. 2023, Espejel et al. 2023, Liu et al. 2023, Huang and Chang 2022, West 2023), but not by the method we propose here.

In Table 3, examples are given of questions that need to be answered using somewhat complex hypothetic-deductive reasoning. All questions and hypotheses are part of what could be called (a theory of) 'basic physics' or elementary 'day-to-day physics', sometimes supplemented with basic biology or basic social psychology (questions Q4, Q10). The table gives the correct answers, ChatGPT's answers, and the main hypotheses that a human reasoner uses, implicitly or explicitly, to deduce the correct answer. The hypotheses are typically laws or regularities, of different degrees of generality. Note that several questions can be *immediately* answered by an adult, without consciously thinking about the hypotheses and without using them in an explicit deductive reasoning. Questions Q1, Q4 and Q10 are examples: one 'sees' the answers immediately; in part also Q6, Q7 and Q9. But even if they are implicit, *the hypotheses and the hypothetic-deductive reasoning are still at the logical basis of the answers*; hence, if pressed to explain or prove answers, one needs to make them explicit. Questions Q2, Q3 and Q5 can only be answered by someone who knows (or who has implicit experience with) some physics. Also here a physicist sees the answers immediately.

| Question | Correct Answer & Main Hypotheses Used | ChatGPT(4)'s Answer | Test passed ? |
|---|---|---|---|
| **Q1.** Suppose five equally sized cubes (1 cm side length) are put in a row next to each other on a wooden table (the cubes touch each other). Cube A is on the left of cube B, which is left of C, which is left of D, which is left of E. Suppose all cubes are made of wood, except cube C, which is made of ice (the room is at -5 degrees Celsius). Suppose I hammer a nail through cube E deep into the table, and heat the room to +20 degrees for | Cubes A and B will move 1 cm to the right, cube C will melt and disappear, and cubes D and E will not move.<br>***Main hypotheses used***:<br>**H1**. Ice melts above 0 degree Celsius.<br>**H2**. When pushed solid objects move in the direction of the pushing force until blocked by some resisting force or object (in the situation depicted).<br>**H3**. Molten ice (water) does not exert resistance to moving solid objects (in the situation depicted). | When you push cube A to the right, it will transmit the force to cube B, and B to C. However, cube C, being made of ice and having been in a room at +20 degrees for an hour, would have melted into water, creating a gap between cubes B and D. Therefore, only cubes A and B will move, and they will move until they reach the position initially occupied by cube C. This movement will be approximately the length of one cube's side, i.e., 1 cm. This is based on the assumptions that the table surface is flat and non-slippery, and that the friction between the cubes | **No** |

| an hour. When I then push Cube A towards the right as far as I can, which cubes will move, and how far? Explain your answer in one to maximum five sentences. | **H4**. Contiguous objects move over the same distance (in the situation depicted).<br>**H5**. Objects fixed by a nail do not move and block the movement of other solid objects (in the situation depicted). | and the table is not sufficient to prevent movement.<br><br><***CORRECT; IRRELEVANT OR WRONG HYPOTHESES***> | |
|---|---|---|---|
| **Q2.** A wire carries a current along the line A-B extending from A to B. Above line A-B a magnetic needle that can rotate is held fixed in a direction along A-B by using wax fixed at its rotation axis. How can I make the needle move by using only a source of heat, e.g. a laser beam? How will the needle move? Explain your answer in one to maximum three sentences. | Heat the wax, and the needle will move until it is perpendicular to A-B.<br><br>***Main hypotheses used***:<br>**H1.** Electrical currents in a wire create magnetic fields perpendicular to the wire.<br>**H2.** Moveable magnet needles (e.g. those that can turn around a rotation axis) move until aligned with nearby magnetic fields, unless a resisting force blocks the movement.<br>**H3.** Wax fixed on the rotation axis of a magnet needle can exert a resisting force (in the sense of H2).<br>**H4.** Heat from a laser beam can melt wax.<br>**H5.** Molten wax fixed on the rotation axis of a magnet needle cannot resist the force on the needle exerted by a current. | You can use the laser beam to heat the wire, which increases its resistance and thus decreases the current flowing through it. This would decrease the magnetic field produced by the current in the wire. If the magnetic field strength decreases enough, it may no longer be sufficient to keep the needle aligned, causing it to move according to Earth's magnetic field or any other residual magnetic fields in the vicinity.<br><br><***INCORRECT; MISSING HYPOTHESES (H3-H5)***> | **No** |
| **Q3.** Two equal vessels, open on the top side, are connected with a long wide tube and filled to the rim with water; both vessels are at the same height. Underneath both vessels is a recipient that contains a chemical substance X. X starts burning when it comes in contact with water. Above the recipient a photodiode is fixed that emits a signal when it detects increasing luminosity. How can one make the diode signal by using only a vertical mechanical force? Explain your answer in one to maximum five sentences. | If one moves one of the vessels horizontally up, water will spill over from the other vessel and fall on X, which will start burning, which will trigger the diode.<br><br>***Main hypotheses used***:<br>**H1.** If among two identical (incl. qua coordinates) and connected vessels fully filled with water one is moved upward, the other vessel will overflow.<br>**H2.** Overflowing vessels spill water to lower regions.<br>**H3.** Chemical substances that start burning increase the luminosity. | Apply the vertical mechanical force downward onto one of the vessels. This will increase the pressure in that vessel, forcing water into the connecting tube and then into the other vessel. As the second vessel overflows, water will spill into the recipient containing substance X. When substance X comes into contact with the water, it will start to burn, emitting light. The photodiode will detect the increased luminosity from the burning substance X and emit a signal.<br><br><***INCORRECT; MISSING HYPOTHESIS (H1)***> | **No** |
| **Q4.** In an aviary a bucket full of water is put in unstable equilibrium on a shelf. Underneath the shelf a small pond contains amphibians. Unfortunately the aviary is kept in a bad condition: the pond contains almost no water anymore, and as a consequence the amphibians are nearing death. How could a large bird in the aviary prolong their life span (with some luck)? How could it end their lives (with some bad luck)? Explain your answer in one to maximum six sentences. | With some luck the bird could push the bucket in unstable equilibrium, so that it falls and thus fills the pond with water. With some bad luck the bucket would fall on the amphibians and kill them.<br><br>***Main hypotheses used***:<br>**H1**. (In the situation depicted) large birds can knock over unstable objects from a shelf, so that the latter fall down.<br>**H2**. Knocked-over buckets containing water can cause small ponds to fill with water (in the situation depicted).<br>**H3**. Amphibians cannot live long without water.<br>**H4**. Prolonging the life span of amphibians can be called 'good luck' (in the situation depicted).<br>**H5**. Falling objects can kill amphibians (in the situation depicted).<br>**H6**. Killing amphibians can be called 'bad luck' (in the situation depicted). | To prolong the amphibians' lifespan, the large bird could, with some luck, fly into or land on the shelf, causing the unstable bucket to topple over and spill water into the nearly dry pond below, replenishing the water the amphibians need to survive. On the other hand, with some bad luck, the bird might only partially displace the bucket, causing the water to spill not into the pond, but elsewhere in the aviary. Additionally, if the bucket topples and lands in the pond, the heavy impact could injure or even kill the amphibians, or the bucket could block the pond, restricting the amphibians' movement and access to the fresh water.<br><br><***CORRECT; LARGELY CORRECT HYPOTHESES***> | **Yes** |
| **Q5.** Suppose an electric circuit contains three elements connected in series: a lamp, a resistance R and a voltage source. Suppose R is at the maximum value allowing for the lamp to emit light at the lowest threshold that can be detected by a photodiode. | R will increase, so the lamp will stop emitting light, so the photodiode will not detect light anymore (the electrical signal of the photodiode will vanish).<br><br>***Main hypotheses used***: | Heating the resistance R would cause its value to increase due to the temperature coefficient of resistivity, which characterizes how a material's electrical resistance changes with temperature.<br>As the resistance in the circuit increases, the current flowing through the circuit decreases (according to Ohm's law: I = V/R), including the | **Yes** |
| What is likely to happen with the photodiode if I heat the resistance? Explain your answer in one to maximum five sentences. | **H1**. The current in serial circuits is inversely proportional to the total resistance (an application of Ohm's law).<br>**H2**. The total resistance of a serial circuit is the sum of the serial resistances.<br>**H3**. Heating resistances (normally) will cause them to increase.<br>**H4**. The luminosity of lamps is proportional to the current going through them (at least in approximation, in the depicted situation).<br>**H5**. Photodiodes not receiving light do not emit a signal (in the depicted situation). | current through the lamp. This will reduce the intensity of light emitted by the lamp, likely to a point below the detection threshold of the photodiode, causing the photodiode to cease signaling the detection of light.<br><br><***CORRECT; CORRECT HYPOTHESES***> | |
| **Q6.** Two identical metal buckets are filled with the same amount of water. The buckets are placed on an induction heater each; the two heat sources deliver identical amounts of heat and power, in such a way that the water levels in both buckets decrease at the same pace, due to evaporation. Suppose ten shirts are lying next to the buckets, and suppose you can move the shirts freely through space in any direction, and to any point by using a gripping system. How can you decrease the level in one bucket considerably by only moving the shirts in horizontal and vertical directions? Explain your answer in one to maximum three sentences. | Plunging the 10 shirts in one of the buckets will absorb quite some water; suffices then to remove the shirts from the bucket.<br><br>***Main hypotheses used***:<br>**H1**. In the situation depicted the shirts can be moved and dipped in one of the two buckets (by using the gripping system).<br>**H2**. In this situation the shirts will absorb water.<br>**H3**. Taking out (by only horizontal and vertical movements) shirts that have absorbed water keeps the water caught in them for some time. | You can use the shirts to form a screen or barrier around one of the buckets, creating an insulated area that prevents the heat generated by the induction heater from escaping as easily. This increases the temperature inside the screen, which increases the rate of evaporation in the bucket within the screen compared to the other bucket. As a result, the water level in the screened bucket decreases faster.<br><br><***INCORRECT; MISSING HYPOTHESES (H1-H3)***> | **No** |
| **Q7.** Suppose Bob pours about a liter of water at room temperature over his cat, applies a small quantity of shampoo on the cat's back, and starts washing his pet. In this process, a typical white substance is formed, as we all know. For a given quantity of shampoo, water, and time, which is/are the main parameter(s) that influence(s) the amount of white substance formed? Can you roughly explain how the process of formation of the white substance works? Explain your answer in one to maximum five sentences. | The main parameters influencing the amount of foam are the intensity with which Bob rubs the hair of the cat, and the extent of the surface over which he rubs. The rubbing increases the surface of interaction between soap and hair and stimulates bubble formation.<br><br>***Main hypotheses used***:<br>**H1**. Shampoo, when rubbed over wet hair, generates foam.<br>**H2**. The harder the rubbing (mentioned in H1), the more foam.<br>**H3**. The larger the hair surface over which one rubs, the more foam.<br>**H4**. Rubbing increases the interaction surface between soap and hair and stimulates bubble formation. | The main parameters that influence the amount of white substance, which is foam or lather, formed are the agitation (or the intensity and speed of washing) and the hardness of the water. The process of forming foam involves trapping air bubbles in a liquid, creating a dispersion of gas in liquid. In the case of washing a cat, the shampoo contains surfactants, which reduce the surface tension of water, allowing it to mix with oils on the cat's fur. Agitation from washing helps to mix air into the water-shampoo mixture, forming bubbles. If the water is hard (high in minerals like calcium and magnesium), it can reduce the effectiveness of the surfactants and thus the amount of foam produced.<br><br><***MOSTLY CORRECT; MOSTLY CORRECT HYPOTHESES***> | **Yes** |
| **Q8.** Consider a plastic cylinder (15 cm diameter, 20 cm height, the upper side is open) whose lower side ends in a funnel, closed at the lower tip. Consider 2 cases, A) and B). In case A), I fill the cylinder with sugar; in case B), I fill the cylinder with water. Suppose now I cut the tip of the funnel so that an opening of about 1 cm diameter is formed. What happens in case A) and in case B) (describe the ensuing process in one sentence) ? What is the shape of the upper surface of the sugar during this process (case A) ? What happens inside the water during this process and how does the water surface look (case B)? Explain | Both sugar and water will flow out of the cylinder. The upper surface of the sugar will be smooth and show a downward cusp above the hole of the funnel (case A). Large bubbles will be formed in the water, and the surface will look irregular (case B).<br><br>***Main hypotheses used***:<br>**H1**. In the situation depicted, water and sugar are attracted by gravity and flow down through the opening, where they do not experience resistance.<br>**H2**. Water removed from the reservoir is replaced by air (from underneath), forming bubbles.<br>**H3**. The bubbles (mentioned in H2) create an irregular surface.<br>**H4**. Sugar removed from the reservoir is also replaced by air (from above), whereby sugar | In case A), the sugar would flow out slowly through the funnel due to its granular nature, while in case B), the water would flow out quickly due to its liquid state and lower friction.<br><br>The upper surface of the sugar would likely form a conical shape as the sugar from the center of the cylinder tends to flow down first due to gravity, which causes the sugar at the edges to slide down and fill the center space, maintaining a slope.<br><br>In case B), the water inside would flow downwards and out of the funnel, and its surface would remain flat due to surface tension and the property of liquids to conform to their container. However, some turbulence may occur near the funnel opening due to the swift motion of the water exiting the funnel. | **No** |

| your answer in one to maximum five sentences. | grains through friction gradually fall into the depression that is created, forming a typical downward cusp (as experience shows). | <***INCORRECT; IRRELEVANT AND MISSING HYPOTHESES (H2-H3)***> | |
|---|---|---|---|
| **Q9.** Alice opens a bottle of a carbonated soft drink, and hears a characteristic hissing sound. She thinks: “That’s a cool sound, how can I increase it?” What is likely the simplest way to increase the sound? What is the key chemical property of substances that, if added in the bottle, would reduce the sound? Explain your answer in one to maximum five sentences. | The simplest way to increase the sound is to vigorously shake the bottle just before opening. The key property of such substances is that they reduce carbon dioxide formation.<br>***Main hypotheses used***:<br>**H1**. Shaking of carbonated soft drinks increases $CO_2$ bubble formation, increasing the pressure in the shaken bottle.<br>**H2**. This pressure can be released by opening the bottle.<br>**H3**. Releasing the pressure due to increased bubble formation (as mentioned in H1-H2) leads to a hissing sound.<br>**H4**. A substance that reduces carbon dioxide formation reduces bubble formation (in the situation depicted in the question). | The simplest way to increase the hissing sound would likely be to increase the pressure in the bottle before opening it, which could be done by shaking the bottle. The hissing sound is produced by the rapid escape of carbon dioxide gas, which is dissolved under pressure in the soft drink; shaking the bottle increases the pressure, causing more gas to be released upon opening.<br>The key chemical property of substances that would reduce the sound if added to the bottle is their ability to reduce the solubility of carbon dioxide in the drink. A substance that increases the pH of the drink (i.e., a base) would cause the carbon dioxide to remain in the solution, reducing the amount of gas released when the bottle is opened and therefore reducing the hissing sound.<br><***OVERALL INCORRECT; IRRELEVANT AND MISSING HYPOTHESES (H4)***> | **No** |
| **Q10.** A leaf is gently stretched by two forces: one force (applied to one end of the leaf) pulls it to the left, the other force (applied to the other end) pulls it to the right. The leaf does not break. Suppose that I put a caterpillar on the leaf, and that after some time, say 15 minutes, the leaf breaks. What is the likely cause? Explain your answer in one to maximum three sentences. | The caterpillar has eaten away some of the leaf, making it weaker, and thus causing the breaking.<br>***Main hypotheses used***:<br>**H1**. Tree leaves may resist stretching forces (in the situation depicted).<br>**H2**. Caterpillars may eat parts of leaves (in the situation depicted), creating holes.<br>**H3**. The holes mentioned in H2 may reduce the strength of leaves. | The likely cause is the added weight and/or the movement of the caterpillar on the leaf. Even though the leaf could withstand the initial two forces stretching it, the additional stress introduced by the caterpillar likely exceeded the leaf's tensile strength, leading to its breakage. The caterpillar's movement might have also caused a change in the distribution of forces on the leaf, contributing to the break.<br><***OVERALL INCORRECT; IRRELEVANT AND MISSING HYPOTHESES (H2-H3)***> | **No** |

**Table 3**. Examples of questions testing ChatGPT(4)’s hypothetic-deductive reasoning. The second column lists correct answers (in our personal judgement) and the main hypotheses (laws, regularities,…) that must be used for answering the questions, explicitly or implicitly. The third column gives ChatGPT(4)’s answers (test done on 24.07.2023). The answers to Q1, Q4, Q5 and Q7 are correct or almost correct; the ‘hypotheses-listing’ test is passed only for Q4, Q5, Q7, as indicated in the last column. In <*brackets*>: first our evaluation of the answer; then an indication of whether the associated hypotheses are correct or not (see Fig. 3 for examples of ChatGPT(4)’s full answer to Q-HYP).

We tried to ask questions that had, a priori, a relatively low likelihood of being discussed in texts (part of ChatGPT’s training set), while still being easily answerable by humans. This appeared to demand some reflection and creativity: ChatGPT is now trained with an enormous set of (web)texts[7]. The main strategy consisted in asking questions with a basic physics content, but probing also knowledge that humans extract from daily experience rather than physics texts: Q1, Q4, Q6, Q7, Q8, Q9 and Q10 are examples. A similar strategy is to ask questions probing reasoning that combines different contexts or theories (Q2, Q4, Q6-Q10).

[7] It is difficult to find precise numbers; one finds numbers ranging from 570 GB to 45 TB or more.

Remarkably, ChatGPT(4) was capable of answering questions Q1, Q4, Q5 and Q7 fully (or almost fully) correctly, even if these answers are based on at least four to six hypotheses. ChatGPT(3) performed less well (test done in April 2023), answering only Q4 and Q5 correctly; this difference is not as salient as for the causal reasoning test. In many of the answers in column 3 of Table 3, ChatGPT(4) *seems* to follow a correct logic; reasoning is simulated or 'emergent'.

Now, our main point is that a better insight in the understanding/reasoning of ChatGPT can quite straightforwardly be obtained: namely by asking it to explicitly list the hypotheses it uses. This appeared to be feasible by simply adding, after ChatGPT had answered a question, a second question (Q-HYP): "Can you, in order to give this answer, reason step by step and list all the hypotheses and laws of nature you use to come to your answer?" That judicious prompting ('chain-of-thought prompting', CoT) can increase LLMs' performance is well-known by now (e.g. Wei et al. 2022, Chen et al. 2023). Since we use one extra question, our method is a variant of the 'zero-shot-CoT' of (Kojima et al. 2022): we explicitly ask to list all hypotheses. Of course, asking Q-HYP is essentially what one would do to test the understanding of humans, who must be able to answer 'why' questions in order to show that they did not arrive at the correct answer by chance or blind copy. Answering such explanation-seeking 'why' questions involves referring to laws, generalities, rules (Hempel 1965) – in general hypotheses, hence our 'hypotheses-listing' test illustrated below. Let us be slightly more precise: the capacity of listing hypotheses is a necessary condition for the capacity of (simulating/emulating) hypothetic-deductive reasoning[8].

With the Q-HYP prompt we could always trigger ChatGPT to provide a list of hypotheses that were at least logically related to the answer it gave. In the case of questions Q4, Q5 and Q7 the given hypotheses corresponded essentially to ours (cf. Table 3); so for these questions the test was passed, as indicated in the table. For the other questions, at least some of the listed hypotheses were irrelevant or wrong, leading to wrong answers. In Figure 3, we illustrate this procedure for questions Q1, Q2 and Q6 of Table 3, and indicate wrong or irrelevant hypotheses.

[8] The second condition is the ability to logically deduce answers from hypotheses. But that ability is more difficult to test.

**ChatGPT(4)'s answers after asking Q-HYP** ("Can you, in order to give this answer, reason step by step and list all the hypotheses and laws of nature you use to come to your answer?").

**Q1:**

Sure, let's break this down step by step:
1) Hypothesis 1: The ice cube (Cube C) will melt when the room temperature is raised to +20 degrees Celsius for an hour.
This is based on the principle of phase transition. Ice melts into water at temperatures above 0 degrees Celsius.
Law of Nature 1: Phase Transition.
Phase transitions are physical changes that occur when matter transitions from one state (solid, liquid, gas) to another due to changes in temperature and pressure. In this case, when the ice cube is exposed to temperatures above its melting point (0 degrees Celsius), it transitions from a solid state to a liquid state.
2) Hypothesis 2: When cube A is pushed, it will transmit the force to cube B, and B to C.
This is based on the principle of force transmission. When a force is applied to an object, it can transmit that force to another object it is in contact with.
Law of Nature 2: Newton's Third Law of Motion.
Newton's Third Law states that for every action, there is an equal and opposite reaction. When cube A is pushed, it pushes against cube B, transmitting the force.
3) Hypothesis 3: After cube C has melted, there will be a gap between cubes B and D, meaning that cubes D and E will not move when cube A is pushed.
**This is based on the assumption that the water resulting from the melted ice cube will not provide enough resistance to transmit the force from cube B to cube D**.
These three hypotheses and corresponding laws of nature lead to the conclusion that when cube A is pushed to the right, only cubes A and B will move, and they will move approximately the length of one cube's side, which is 1 cm.

**Q2:**

Sure, here are the steps, each accompanied by the respective physical principle or hypothesis:
1. Principle of the laser's operation and interaction with matter: When the laser beam hits the wire, it heats it. The laser light energy is absorbed by the wire, increasing its temperature. This is due to the principle that light (in this case, from the laser) can be absorbed by matter, leading to an increase in the internal energy (and thus the temperature) of the material.
2. Ohm's Law and effect of temperature on resistance: As the wire's temperature increases, its electrical resistance increases. This is due to the principle that the resistance of a conductor is directly proportional to its temperature. The resulting decrease in the current is given by Ohm's law, which states that the current through a conductor between two points is directly proportional to the voltage across the two points, and inversely proportional to the resistance between them.
3. Biot-Savart Law or Ampère's circuital law: The magnetic field around a current-carrying wire decreases as the current through the wire decreases. This is given by Biot-Savart Law or Ampère's circuital law, which state that the magnetic field in space around an electric current is proportional to the electric current which serves as its source.
4. Magnetic compass needle alignment: The needle of a magnetic compass aligns with the magnetic field lines. In this case, the magnetic field is primarily produced by the current in the wire, so if the magnetic field due to the wire decreases, the needle may no longer align with it.
5. Earth's magnetic field: If the magnetic field produced by the wire decreases enough, the magnetic needle may align with the Earth's magnetic field, or any other residual magnetic fields in the vicinity. This is because a magnetic needle, when free to rotate, tends to align with the net magnetic field at its location.
These steps lead to the conclusion that the needle will move upon heating of the wire by the laser. The specific movement of the needle would depend on the relative strengths and directions of the magnetic field produced by the wire and the Earth's magnetic field or other nearby magnetic fields.

**Q6:**

Sure, let's break this down:
1. Law of Thermodynamics (Heat Transfer): Heat transfer occurs from a region of high temperature to a region of low temperature. If we isolate one of the buckets with the shirts, we're creating a barrier that keeps more of the heat around that bucket. As a result, the temperature around the bucket inside the barrier will be higher compared to the one without the barrier.
2. Assumption (Heat Conduction): We're assuming that the shirts are not perfect heat conductors, i.e., they can partially trap the heat within the enclosure. The more shirts we use, the better the insulation effect.
3. Evaporation (Phase Transition): Evaporation is a phase transition from the liquid phase to vapor phase. It is a process that requires energy (heat), and it's temperature-dependent. The higher the temperature, the more molecules have enough energy to overcome the intermolecular forces and change from liquid to gas. Hence, when we increase the temperature around one bucket, we increase the evaporation rate and thus the water level decreases faster.
4. Assumption (Consistent Heat Supply): We assume that the induction heaters are providing a consistent and equal amount of heat to both buckets. This means that any difference in evaporation rates (and thus water levels) between the two buckets is due to the impact of the shirts and not fluctuations in the heat supplied.
5. Assumption (Equal Evaporation at Start): The premise mentions that both buckets start with the same amount of water and are evaporating at the same rate. This sets a baseline for the comparison.
6. Mechanical Work (Shirt Movement): We are assuming that the only actions allowed in the system are the movements of the shirts. We can move the shirts around, but we can't otherwise interact with the buckets, heaters, or surrounding environment.
By applying these laws of nature and assumptions, we can conclude that insulating one bucket with shirts will make the water level in that bucket decrease faster.

**Fig. 3**. Transcription of dialogues in which ChatGPT(4) is prompted to list the hypotheses associated with the answers to Q1, Q2, Q6 of Table 3. First the initial question is asked, then, after ChatGPT(4)'s answer, we asked Q-HYP. Q1 was answered correctly but the listed hypotheses contained irrelevant (underlined) or wrong (bold) hypotheses. For Q2 and Q6 irrelevant hypotheses were given, leading to a wrong answer (compare to Table 3).

Inspection of Table 3 and Fig. 3 shows that ChatGPT(4) is often capable of giving answers that are almost correct even if wrong in principle (Q6, Q8, Q9), or implausible (Q10). This surely is a remarkable achievement. Especially, ChatGPT(4) can associate these (incorrect) answers with *logically related* (incorrect/irrelevant) hypotheses. This is remarkable, since this points to the conclusion that ChatGPT(4) simulates logical reasoning, even if we must assume it is an LLM that finds answers by statistical word correlations. Now, correct answers are not always associated with correct hypotheses; it is here that our hypotheses-listing test is most interesting: see e.g. Q1 in Fig. 3. Humans who would give the correct answer to Q1 but would then explain it by listing ChatGPT(4)'s reasons, would not be credited with understanding neither with correct reasoning.

Based on experiments as the above, as well as other published results e.g. from (Bang et al. 2023, Espejel et al. 2023, Liu et al. 2023, Huang and Chang 2022, West 2023), we conclude that, for the time being, ChatGPT has a limited capacity for (simulating) hypothetic-deductive reasoning. When the questions/context is somewhat sophisticated, notably when these involve combined knowledge fields,

ChatGPT regularly associates irrelevant hypotheses to its (correct or incorrect) answers. For minimally trained humans, let alone experts, these questions should pose less problems.

The method of 'hypotheses listing' by prompting with Q-HYP can, in our experience, be used for analyzing hypothetic-deductive reasoning skills for a wide range of problem types. Of course, a much larger body of evidence could be compiled along the lines of the above test. But we do not intend here to provide a benchmark test but to motivate and propose a testing method. Also, we expect that the above questions, easily answerable by experts, will surely soon be answered better by upgraded models.

## 5. Discussion; significance in the context of AGI.

Within the context of research on AGI, the significance of the analysis presented here hinges on the significance one attributes to hypothetic-deductive reasoning. It is argued elsewhere that *any* form of rationality or thinking is ultimately based on hypothetic-deductive reasoning, implicitly or explicitly (Vervoort et al. 2021). While this may be acceptable in a scientific context – for instance, mathematical/psychological/physical reasoning can be construed as reasoning using the relevant mathematical/psychological/physical laws or theories –, it may seem a strong claim to some when applied to 'day-to-day' reasoning[9]. But, on the formal or epistemic level, common-sense thinking is, arguably, not very different from scientific thinking. It may well be only different-in-degree; sure, it demands concepts and hypotheses that are less sophisticated. In other words, the semantics may be simpler, but the logical backbone, the fundamental syntax, may well be identical. Indeed, from a neurobiological perspective, so from the perspective of the detailed working of the brain when it reasons, it would be surprising that fundamentally different neurobiological mechanisms would be at work for coherent thinking in different semantic contexts. Take mundane thoughts as 'I will buy flowers for Anna', 'I will wear my red shirt, not my green one' or 'The neighbour's dog just bit me in the leg'. They are all part of an (implicit or explicit) hypothetic-deductive scheme, even if the hypotheses involved might be trivial and therefore surely implicit. But these hypotheses need to be made explicit if someone asks 'why?'. 'I will buy flowers for Anna' is a thought that can, for instance, be the logical consequence of 'I want to make a good impression on Anna' and the hypothesis 'Giving flowers usually leaves a good impression' (a regularity, a generalisation). The decision 'I will wear my red shirt, not my green one' might be the logical outcome of ‘Yesterday I wore my green shirt’ and ‘Having some variation in outfit is nice’ (a regularity). ‘The neighbour’s dog just bit me in the leg’ might be the linguistic representation (or rationalisation) of an *immediate* – i.e. *not* mediated by explicit, conscious hypothetic-deductive reasoning – brain process that can yet be rationalised as the logical combination of hypotheses as ‘I saw a big brown creature jumping at me’, ‘at night the neighbour’s dog is loose and nasty’, ‘this type of nasty dog likes to bite’ (a generalisation), and ‘I felt a pain in my leg that might well be due to teeth’. Needless to say, deductions as the above, especially in the case of undeniable ‘truths’, are automatic *in the human brain*; they often do not need to be consciously thought about via a hypothetic-deductive process in order to exist in the brain. In that sense they are unanalyzed and immediate.

[9] We also do not claim that all forms of reasoning can be subsumed under the hypothetic-deductive scheme. But we do believe it is not trivial to find examples of such exceptions. If 'correct reasoning' exists, there must be rules to distinguish correct from incorrect reasoning. Ultimately, these rules might be (included in) the hypotheses of hypothetic-deductive reasoning. Subsuming answers under correct rules (laws etc.) seems what correct reasoning is all about.

But they might still exist in the brain in some form or other – say, in the form of some connected neural nets each corresponding to the different hypotheses. More to the point, *things may be different for machine reasoning*. For machines and robots it may well be necessary to base their functioning and reasoning on explicit, formalised hypothetic-deductive steps – especially if the machine outcomes (decisions, actions, recommendations, answers,…) need to be justified or explained.

As another illustration of the relevance of hypothetic-deductive reasoning for machine functioning and AGI, consider social robots. In a social context such a robot will have to behave according to certain social habits, rules, regularities. (Surely these rules or regularities will at least partly be imposed ‘from outside’ by humans, if only to avoid worst-case scenarios in which advanced robots develop singularly deleterious goals.) The same holds for ethical robots – which could, as an example, be governed ‘explicitly’ by versions of Asimov’s laws of robotics. In the sense that such robots will have to behave according to certain rules and regularities, they will have to master (elementary or advanced) social and/or ethical theories, implicitly or explicitly; in other words, they will have to be capable of ethical and social hypothetic-deductive reasoning.

Now, one might object that LLMs as ChatGPT are already capable of *implicit* hypothetic-deductive reasoning, or at least of emulating such reasoning (which is good enough). They often produce a correct outcome even if they cannot list the correct hypotheses: see examples above; examples as these can be multiplied at will. True, but again, for high-stakes applications genuine, explicit hypothetic-deductive reasoning (or its emulation) seems imperative. As long as the hypothetic-deductive reasoning is implicit, trust in the AI may not be warranted. The only guarantee for reasonably flawless, justifiable, and explainable AGI is therefore, we submit, its capability of explicit hypothetic-deductive reasoning. As argued above, a minimal necessary condition is that ‘reasons’ be transparent; that hypotheses be listable.

## 6. Conclusion.

Following (Vervoort et al. 2021), we have started here from the idea that hypothetic-deductive reasoning is not only a scientific type of thinking, but can subsume a wide range of other types of reasoning. A necessary condition for the capacity of explicit hypothetic-deductive reasoning is the capacity to list the hypotheses that generate an answer, a solution, or a decision. Thus, we proposed a hypotheses-listing test to assess LLMs' and other AI's reasoning skills. We also argued that causal reasoning is a simplified derivative of hypothetic-deductive thinking, and proposed a test based on the compendium from (Paul and Hall 2013). We applied these tests to ChatGPT(3) and ChatGPT(4), in order to illustrate our methods. Based on the results we concluded that, when contexts/questions/knowledge-fields become somewhat sophisticated, these LLMs do not reach expert-level for both types of reasoning. At the same time, since ChatGPT(4) is already considerably more powerful than ChatGPT(3), an evolution to more advanced levels seems highly likely.

In that respect, it is important to have simple criteria for qualifying an AI as AGI. The logical criterion that immediately follows from our model of reasoning is the following: a necessary condition for AGI is the capacity to emulate explicit hypothetic-deductive reasoning *at expert level*, in a sufficiently wide range of problem-spaces or knowledge-fields. We hope our tests may be helpful in this assessment. (Our verdict of whether ChatGPT can already be considered an AGI may be compared to the more optimistic verdict of (Bubeck et al. 2023).) Finally, we hope to have provided a pragmatic answer to the question: "how much more there is to true understanding than on-the-fly improvisation" (Bubeck et al. 2023).

## Appendix A. ChatGPT(4) tested with 26 causal reasoning cases.

In the following Table we summarise the results of our causal reasoning test (see explanations in Section 3).

| Diagram | Correct answer | | Diagram | Correct answer |
| --- | --- | --- | --- | --- |
| **1. (From PH, Fig. 1, p. ix)** | Yes.<br>**C+(t1); D+(t2)** | | **14.** | Yes.<br>**A+(t1), C+(t1); D+(t2), A1+(t2); F-(t3); A2+(t3)** |
| | | | | |
| **(2.)** | Yes.<br>**C+(t1); D+(t2), B-(t2)** | | **15. (From PH, Fig. 42, p. xx)** | Yes.<br>**A+(t1), C+(t1); etc.** |
| **3.** | No.<br>**C-(t1); D-(t2)** | | **16.** | Yes.<br>**A+(t1), G+(t1); A2+(t3), F+(t3) etc.** |
| **4. (From PH, Fig. 2, p. ix)** | Yes.<br>**A+(t1); B+(t2)** | | **17. (From PH, Fig. 43, p. xxi)** | No.<br>**C+(t1); D+(t2); E+(t3)** |
| **5. (From PH, Fig. 10, p. xi)** | Yes.<br>**A+(t1); D+(t2); F-(t3)** | | **18. (From PH, Fig. 44, p. xxi)** | Yes.<br>**H+(t1), C+(t1); H3+(t4), G-(t4); etc.** |
| **6.** | No.<br>**A-(t1) etc.** | | **19.** | No.<br>**F+(t1), C-(t1), A-(t1); G+(t4); etc.** |
| **7.** | No. | | **20.** | No.<br>**C+(t1); F+(t3); etc.** |
| | A-(t1); D+(t2); F-(t3); A2-(t3) | | | |
| **8.** **(From PH, Fig. 12, p. xii)** | Yes.<br>**C+(t1); D+(t2)** | | **(21.)** | Yes.<br>**A+(t1), G-(t1), C-(t1); A2+(t3), F+(t3); etc.** |
| **9.** | No.<br>**C+(t1); B-(t2), G-(t2)** | | **22.** | No.<br>**H2-(t3); H3-(t4)** |
| **10.** **(From PH, Fig. 14, p. xiii)** | Yes.<br>**A+(t1), C+(t1); B+(t2), F+(t2)** | | **23.** | Yes.<br>**F+(t1), C+(t1); F1+(t2); E+(t3); G-(t4)** |
| **11.** | No.<br>**A-(t1)** | | **24.** | Yes.<br>**F+(t1) and effects** |
| **12.** **(From PH, Fig. 29, p. xvii)** | Yes.<br>**A+(t1); C+(t1)** | | **25.** | **Yes.**<br>**F+(t1); F1+(t2); H2+(t3), E+(t3); H3+(t4), G-(t4)** |

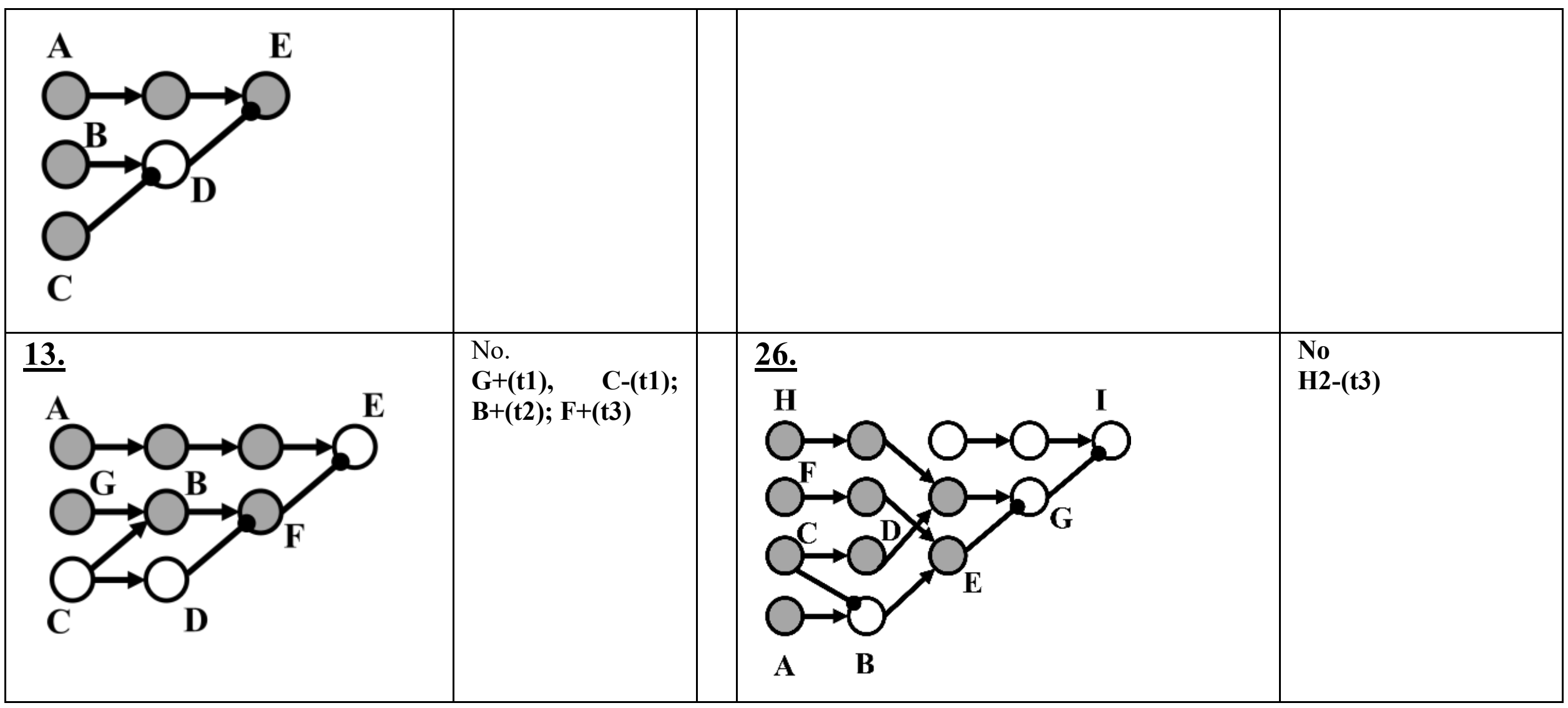

| | | | | |
|---|---|---|---|---|
| | | | | |
| **13.** | No.<br>G+(t1), C-(t1);<br>B+(t2); F+(t3) | | **26.** | No<br>H2-(t3) |

**Table 1**. All diagrams of the causal test, in which we asked ChatGPT(4) whether the last neuron in the diagram fires or not, and what the causes of the firing or not firing are. In column 2 and 4 the correct answers are given (when no confusion exists we do not always mention all causes). For examples of ChatGPT(4)'s full answers, see Table 2. Here we only indicate whether ChatGPT(4) gave incomplete or wrong answers by underlining the diagram number, as in diagram **3**, **5**, etc. Partially correct answers are indicated with extra brackets, as in **(2)**. Many diagrams are discussed in (Paul and Hall 2013) (PH in the table).

## Appendix B. Analytic definition of the concept of cause in neuron diagrams.

For all the cases of Table 1, we identified the causes X of event Y (Y is the firing or non-firing of the last neuron in the diagrams, usually labelled E) by following definition, applied to a given neuron diagram:

X is a cause of Y *iff* ¬X (ceteris paribus, off-path under max blocking) leads to ¬Y. (DEF1)

Here '¬X' can be read 'not-X' (event X not happening, i.e. inverting 'on' and 'off' states of the X-neuron). (X and Y stand in principle for the events, firing or non-firing, but sometimes also indicate the neurons. If confusion is possible, we write e.g. 'X-neuron'.) The right-hand side needs some explanation; DEF1 comes with a rule. To see whether X is a cause of Y, one needs to consider the factual scenario as depicted in the diagrams, and a counterfactual scenario that corresponds to the diagram in which X is replaced by ¬X, and the consequences are implemented ceteris paribus, so while keeping all other events, not related to X, fixed. If two or more paths start from a 'bifurcating' X, one also needs to distinguish the 'direct' path from X to Y (containing 'in-path' events); and the 'indirect' path(s) from X to Y (containing 'off-path’ events); all events that are not in-path in the direct path are off-path. Indeed, in many diagrams neurons are connected to more than one neuron; e.g. in Fig. 1 two paths run from C to E. The direct path is, by definition, the shortest path, once all ‘redundant’ neurons are collapsed on their antecedent ‘parent’ with the same on/off state; such redundant neurons have, by definition, one connection in and out. For instance, D in Fig. 1 is a redundant neuron in this sense; and long chains of identical neurons can thus be made shorter.

With these definitions we can make DEF1 explicit. This is most straightforward if X is non-bifurcating, so if there is one path from X to Y, *or* if the X-neuron is off (whether bifurcating or not). Then we do not even need the ‘off-path under max blocking’ clause and our rule or definition corresponds to the simple counterfactual method. Consider the case that X is off and bifurcating. Then DEF1 just stipulates that event X (the X-neuron being off) is a cause of Y just in case ¬X implemented (in the counterfactual diagram), ceteris paribus, leads to ¬Y. This case is illustrated in Fig. 2 (case 3 in Table 1), where two paths run from C to E.

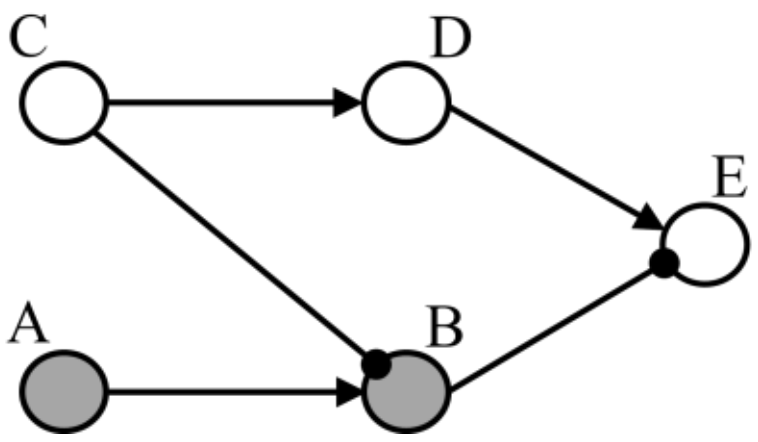

Fig. 2. The non-firing of C is a cause of the non-firing of E.

One intuitively sees that here the non-firing of C is a cause of the non-firing of E: if C would fire, E would fire – because the counterfactually firing C would block the B-neuron. This is also what DEF1 delivers. Note that this is a (counterfactually) ‘off-path maximal blocking’ scheme, in the sense that in the counterfactual diagram the off-path B is blocked from firing. Off-path blocking is maximal in that it is implemented in the counterfactual scheme.

If the X-neuron under scrutiny is ‘on’ and bifurcating, the situation is slightly more complex: now we explicitly need the ‘off-path maximal blocking’ clause: we need to implement ¬X while *maintaining* X’s blocking off-path, i.e. in any segment of all indirect paths. This allows for instance for identifying the firing of C as a cause of the firing of E in Fig. 1, as one would spontaneously do. Thus, DEF1 now stipulates, more precisely, to implement ¬X in the direct path and in all indirect paths, while *maintaining* off-path (so in any indirect path) any blocking that can be retraced to the (factually firing) X. This is somewhat sophisticated, because it boils down to not fully implementing ¬X; off-path one needs to maintain any blocking in the indirect path that can be retraced to the factual firing of X. (This definition also works if the blocking occurs further downward the root neuron, as one can easily verify.) In sum, in case the X-event under scrutiny corresponds to a bifurcating and firing X-neuron, we now explicitly need the 'off-path under max blocking' clause. The blocking in the indirect path is now maximal in that it is maintained from the factual firing of X. (Recall, if X is off, off-path blocking is maximal in that it is *implemented* in the counterfactual scheme.)

With these specifications, DEF1 leads in all diagrams of Table 1 to causes that correspond to intuitive causes.